\documentclass[a4paper, 10pt, conference]{ieeeconf}      
\usepackage{graphicx}
\usepackage{amssymb}
\usepackage{xfrac}
\usepackage{FG2024}
\usepackage{amsmath}
\usepackage{adjustbox}
\usepackage{afterpage}
\usepackage{makecell}
\FGfinalcopy 
\usepackage[pagebackref=true,breaklinks=true,letterpaper=true,colorlinks,bookmarks=false]{hyperref}

\newcommand{\TY}[1]{\textcolor{black}{#1}}

\IEEEoverridecommandlockouts                              
\def\FGPaperID{96} 

\title{\LARGE \bf
Embedded Representation Learning Network for Animating Styled Video Portrait}

\author{\parbox{16cm}{\centering
    {\large Tianyong Wang,  Xiangyu Liang,  Wangguandong Zheng, Dan Niu, Haifeng Xia and Siyu Xia}\\
    {\normalsize
     School of Automation, Southeast University, Nanjing, China}}
    \thanks{This work was supported in part by Key Laboratory of Intelligent Processing Technology for Digital Music (Zhejiang Conservatory of Music), Ministry of Culture and Tourism under Grant 2023DMKLB003.}
}

\usepackage{fancyhdr}
\begin{document}


\ifFGfinal
\thispagestyle{empty}
\pagestyle{empty}
\else
\pagestyle{empty}
\fi
\maketitle

\thispagestyle{fancy}

\begin{abstract}
The 
talking head generation recently attracted considerable attention due to its widespread application prospects, especially for digital avatars and 3D animation design. Inspired by this practical demand, several works explored Neural Radiance Fields (NeRF) to synthesize the talking heads. 
\TY{However, these methods based on NeRF face two challenges: (1) Difficulty in generating style-controllable talking heads. (2) Displacement artifacts around the neck in rendered images. 
To overcome these two challenges,}
we propose a novel generative paradigm \textit{Embedded Representation Learning Network} (ERLNet) with two learning stages. First, the \textit{ audio-driven FLAME} (ADF) module is constructed to produce facial expression and head pose sequences synchronized with content audio and style video. Second, given the sequence deduced by the ADF, one novel \textit{dual-branch fusion NeRF} (DBF-NeRF) explores these contents to render the final images. Extensive empirical studies demonstrate that the collaboration of these two stages effectively facilitates our method to render a more realistic talking head than the existing algorithms.


\end{abstract}

\section{INTRODUCTION}
Speech-driven talking head generation is a technology with numerous applications, such as digital avatars and video dubbing. However, producing high-fidelity talking head videos is still a complex task. By now, researchers \cite{wav2lip, audio2head, au, dinet} have explored the generation of talking head videos in the 2D image space. Furthermore, other studies \cite{flow-guided, facial-synthesizing} used prior 3D knowledge to enhance the information provided to the network. In real-world scenarios, people show various facial expressions and head poses while speaking. To approximate this practical demand, there have been efforts \cite{evp, eamm, pcavs, emotion-wav2lip, gcavt} to generate style-controllable talking head videos.


Recently, Neural Radiance Fields (NeRF) \cite{nerf} have demonstrated remarkable representational capabilities in the context of 3D scenes. 
Guo et al. \cite{adnerf} presented AD-NeRF, the first NeRF-based method for generating talking head videos. They used separate NeRFs to render the head and torso, but this approach easily introduces potential artifacts in the neck region, especially during extensive head movement. Others such as DFRF \cite{dfrf} and LipNeRF \cite{lipnerf} treat the torso portions of the original image as a background and render the head on this background. These approaches need corresponding torso images for rendering different head poses. On the other hand, SSP-NeRF \cite{sspnerf} takes a different approach by incorporating a global deform module, which can affect the rendering results of the facial region.


From the aforementioned NeRF-based means, it is evident that these algorithms have limitations when it comes to effectively generating style-controllable talking head videos. They also face challenges in accurately rendering both the head and the torso. Therefore, the application of NeRF to the generation of style-controllable talking head videos remains several under-explored challenges:
1) Converting the head pose into a camera parameter can enhance the rendering quality of the head. However, the movement patterns of the torso differ from those of the head. Thus, it is important to maintain the consistency of movement between head and torso for the generation of talking head video.
2) Due to the scarcity of style content, it becomes difficult to generate the style-controllable talking head video with these existing datasets. Moreover, they rarely focus on capturing the movement of the torso, such as the MEAD \cite{mead} dataset.

\begin{figure}[t]
  \centering
   \includegraphics[width=1.0\linewidth]{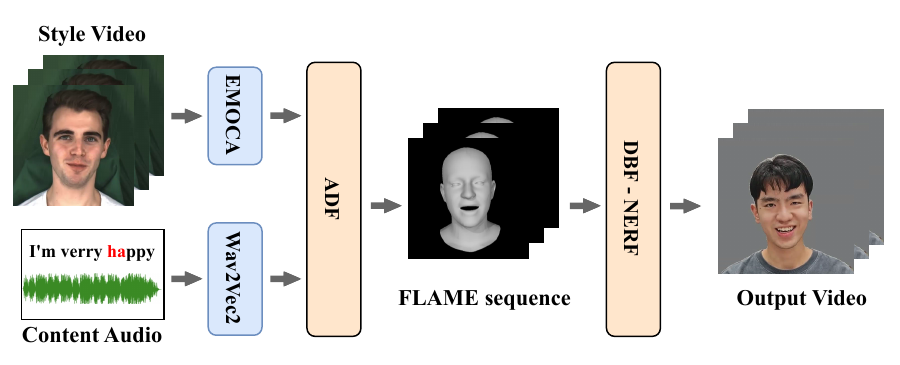}
   \caption{Overview of ERLNet. EMOCA \cite{EMOCA} and Wav2Vec2 \cite{wav2vec2} are employed for extracting FLAME coefficients and audio features. During the inference process, our approach requires two inputs: a style reference video and a content audio. After passing through the ADF module, a FLAME coefficients sequence matched to the speech is obtained. Subsequently, the DBF-NeRF network takes these FLAME coefficients as input to render the final video.}
   \label{fig:overview}
\end{figure}

In this paper, we introduce \textit{Embedded Representation Learning Network} (ERLNet), a style-controllable talking head video generation method based on NeRF. \TY{Here, the style encompasses: (1) facial expressions and (2) head movements.}
As shown in Fig.\ref{fig:overview}, style videos and content audio are used as input, with Faces Learned with an Articulated Model and Expressions (FLAME) \cite{flame} coefficients as an intermediate representation.
\textbf{First}, we design and deploy the \textit{audio driven FLAME} (ADF) module based on VQ-VAE to establish the mapping between the speech, style video, and FLAME coefficients. Specially, 
Our method decouples expression and head pose, allowing us to achieve more precise control over style. \textbf{Second}, considering the difference between head and torso movements, the other \textit{dual-branch fusion NeRF} (DBF-NeRF) model is built to enhance the realism of the generated results. In contrast to existing NeRF-based methods, 
we employ FLAME coefficients as input conditions for NeRF. This approach allows us to control the poses and facial expressions of the rendered video. \textbf{Finally}, we have collected a \textit{long-duration styled talking video dataset} (LDST). In LDST, each video segment includes content of at least 30 seconds and involves five distinct facial expressions along with head movements. Importantly, there are minimal torso movements between different video segments, which facilitates NeRF training. The contributions of our work are summarized as follows:
\begin{itemize}
    \item We propose a novel algorithm \textit{Embedded Representation Learning Network} to generate style-controlled talking heads. Compared to existing methods, our ERLNet synthesizes more realistic talking head videos.
    \item To achieve the controllable generative procedure, we design \textit{audio driven FLAME} module to capture the relation between speaking style, facial expression, and head movement and introduce a \textit{dual-branch fusion NeRF} model to address the issue of artifacts in the neck region.
    \item Inspired by the significant defects of the existing benchmarks, this work collected and built a \textit{long-duration talking head video dataset} and each video segment consists of five different facial expressions.
\end{itemize}

\section{RELATED WORK}

\subsection{Speech Driven Talking Head}

\subsubsection{2D based talking head}

Wav2Lip \cite{wav2lip}  proposed a CNN-based generative network and used a pre-trained discriminator to enhance the quality of the generated lip movements.
Audio2Head \cite{audio2head} introduced an image motion field to model facial motion and used LSTM \cite{xia2022adversarial} to predict the relationship between speech and head pose.
Chen et al. \cite{au} utilized information from audio and action units as driving factors.
Ravichandra et al. \cite{cross-model-disentanglement} employed a specialized encoder-decoder framework in which two decoders were used to process the mouth and face separately, and they were fused at the final feature layer.
DINet \cite{dinet} designed a feature map-based deformation network to preserve more details of the oral cavity.
In recent years, diffusion models have garnered significant attention in the field of image generation, and some studies \cite{difftalk, audio-condition-diffusion} use them to generate talking heads.


\subsubsection{3D based talking head}
Due to their inability to take into account depth information, 2D-based generation methods often produce videos lacking realism. Therefore, some work focuses on utilizing 3D prior or depth information to generate a talking head. Several studies \cite{voicemesh, faceformer, codetalker, lipsync3d, meshtalk} focus on the recovery of 3D facial animations from speech, resulting in controller parameters for 3D models or sets of facial mesh vertex coordinates. Additionally, some other studies \cite{flow-guided, facial-synthesizing} treat 3D Morphable Face Model (3DMM) coefficients as a form of prior knowledge, serving as an intermediate representation within the network.


In recent years, NeRF has achieved significant success in the representation of 3D scenes, eschewing the use of meshes or point clouds in favor of an implicit neural radiance field. Several studies \cite{adnerf, dfanerf, dfrf, lipnerf, sspnerf} have effectively applied the NeRF model to talking head generation tasks.
AD-NeRF \cite{adnerf} is the first work to apply NeRF in the field of talking head generation. It achieves this by separately modeling the head and torso using two independent NeRFs.
LipNeRF \cite{lipnerf} does not take audio features as input; instead, it discusses the advantages and disadvantages of using 3DMM coefficients as input.
DFRF \cite{dfrf} accelerates NeRF training by designing a Face Warping Module to effectively leverage information from each frame of the image sequence.
DFA-NeRF \cite{dfanerf} designed a Transformer-based VAE to predict head poses and blinking.
SSP-NeRF \cite{sspnerf} uses a semantic-aware module to adaptively select sampling points. In addition, it incorporates a deformation module to address inconsistencies in head and torso motion.

\subsection{Style-controllable Talking Head}
However, the methods mentioned above only address the generation of talking heads with audio as input conditions. In practice, speakers can employ different speaking styles. Generating style-controllable talking heads remains a challenging task.

EVP \cite{evp} uses landmarks, 3DMM coefficients, and edge maps as intermediate representations. It utilizes expression encoding as an additional input.
Wu et al. \cite{imitating-arbitrary} employ a specific formula to extract the speaking style.
Ye et al. \cite{dynamic-neural-textures} proposed dynamic neural textures to generate emotion-controllable talking head.
Goyal et al. \cite{emotion-wav2lip} proposed a framework similar to Wav2Lip. They mask the entire image and apply an emotion discriminator.
EAMM \cite{eamm} employs a facial-dynamics module based on key points and a dense warping field.
GC-AVT \cite{gcavt} decouples facial attributes using image enhancement techniques.
PC-AVS \cite{pcavs} employs a pose-controllable method without using structural intermediate information.


\section{PROPOSED METHOD}

\subsection{Problem Setup \& Motivation}
In recent years, neural radiance fields (NeRF) achieves promising performance on generating static scenes by capturing implicit information from multiple viewpoint images \cite{nerf}. 
Since NeRF constructs a continuous volumetric radiance field, it successfully renders a 2D image in a 3D space. However, NeRF and its variants \cite{mipnerf, blocknerf} hardly describe the association of multiple modalities during the rendering procedure. \textit{For example}, when performing speech-driven talking head generation, 
NeRF only renders a 3D scene per frame and ignores the match between the expression/pose of the host and voice. This drawback easily results in the obvious difference between the real video and the rendered one. Thus, generating high-fidelity talking head video becomes a challenging task.


Motivated by the practical demand, this paper proposes a novel learning algorithm \textit{Embedded Representation Learning Network}
 (ERLNet) to animate video portraits by considering multi-modality accordance with two stages. 
First, \textit{audio driven FLAME} (ADF) module aims to learn the latent feature of facial expression and head pose and store the intrinsic representation in two independent codebooks under the VQ-VAE framework. 
Then,
content semantics is extracted from the given audio and combined with style features of pose and expression to acquire the stylized FLAME coefficients sequence.
For the second stage, in \textit{dual-branch fusion NeRF} (DBF-NeRF), Head-NeRF and Static-NeRF consider FLAME coefficients as input and render the output video collaboratively.



\subsection{Audio Driven FLAME}
Animating video portraits aims to stylize talking head video with the given audio and achieves the match of multi-modality, e.g., facial expression, head pose, and audio. To overcome this challenge, the fusion and editing of features cast light in the dark.



Along with this direction, the first step is to learn latent representations of facial expressions and head poses. Inspired by the successful application of VQ-VAE \cite{codetalker}, we
build the FLAME latent space learning a VQ-VAE codebook. In the second step, two independent AutoEncoders (AE) \cite{xia2020embedded,xia2020hgnet} map the expression sequence, pose sequence, and audio into the FLAME latent space and generate FLAME coefficients sequence through the frozen decoders.

\subsubsection{FLAME latent space} 
\begin{figure}[t]
  \centering
   \includegraphics[width=0.9\linewidth]{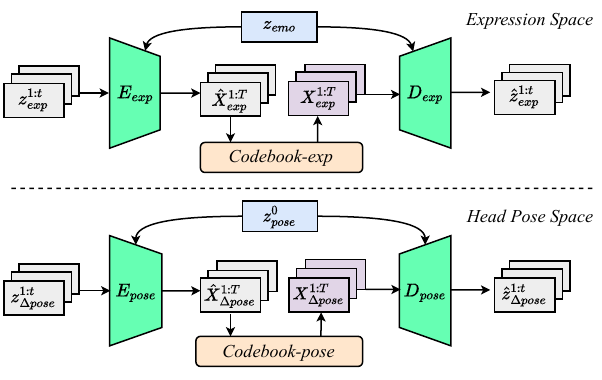}
   \caption{Structure of FLAME latent space. \TY{Expression sequence $z_{exp}^{1:t}$ 
   and head pose variation sequence $z_{\Delta pose}^{1:t}$ of length $t$ are inputs.}
   Two individual VQ-VAEs are used to build the expression space and the \TY{head} pose space. \TY{Emotion vector $z_{emo}$ and the head pose of the first frame $z_{pose}^0$} are utilized as conditional inputs. Only decoders and codebooks are \TY{frozen and} used in subsequent networks.}
   \label{fig:latentspace}
\end{figure}

\begin{figure}[t]
  \centering
   \includegraphics[width=1.0\linewidth]{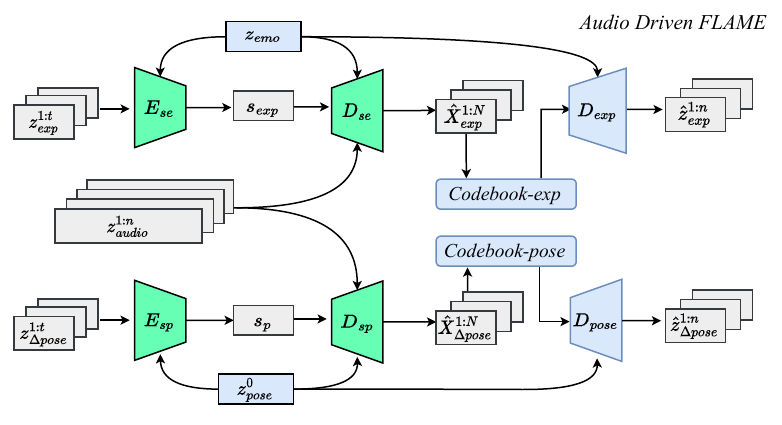}
   \caption{Structure of ADF. \TY{With expression sequence $z_{exp}^{1:t} $ and head pose variation sequence $z_{\Delta pose}^{1:t}$ of length $t$ as inputs, two encoders $E_{se}$ and $E_{sp}$ are employed to extract expression style feature $s_{exp}$ and head pose style feature $s_p$. Then, audio feature $z_{audio}^{1:n}$ of length $n$, $s_{exp}$ and $s_p$  are combined. Two decoders $D_{se}$ and $D_{sp}$ map them to the pre-trained codebooks. After that, pre-trained decoders $D_{exp}$ and $D_{pose}$ are employed to generate the final expressions and poses. Emotion vector and the initial pose are utilized as conditional inputs.}}
   \label{fig:a2f}
\end{figure}

\TY{Previous work \cite{codetalker} built a VQ-VAE discrete codebook to promote the motion synthesis quality against cross-modal ambiguity.}
\TY{Inspired by this, we }
employ the FLAME head model as prior knowledge to establish both the expression space and the head pose space. 
\TY{As shown in Fig.\ref{fig:latentspace}}, 
we employ two individual transformer-based conditional VQ-VAEs to model the expression space and head pose space. Each VQ-VAE consists of three main components: an encoder, a decoder, and a codebook. 

For expression space, the encoder takes the expression sequence \TY{$z_{exp}^{1:t} \in \mathbb R^{t\times 53}$ of length $t$} as input. 
\TY{Emotion vector $z_{emo} \in \mathbb R^5$ is used as a conditional input to both the encoder and decoder in the form of one-hot encoding.}

For head pose space, rather than directly reconstructing the pose, we focus on reconstructing the variation of pose \TY{$z_{\Delta pose}^{1:t} \in \mathbb R^{t\times 6}$},
\TY{which contributes to a more stable and realistic outcome.}
The pose of the first frame \TY{$z_{pose}^0 \in \mathbb R^6$} is used as a conditional input.


\TY{Expression space and head pose space follow analogous procedures.} 
The input sequence is first embedded as a latent code sequence by transformer-based encoder:
\begin{equation}
\TY{
    \hat X^{1:T} = E( z^{1:t}),}
\end{equation}
\TY{where $\hat X^{1:T}$ is the latent code sequence of length $T$, and $T = \tau t$, $\tau$ is the scaling factor. $E$ is the encoder ($E_{exp}$ or $E_{pose}$) function, and $z^{1:t}$ is the input sequence of length $t$.} Then, we employ an element-wise quantization function to map each latent code to its closest codebook $C$ entry through:
\begin{equation}
\TY{ X^{1:T} = Q(\hat X^{1:T}):=\mathop{\arg\min}\limits_{c^k \in C}\|\hat x^i - c^k\|_2.}
\end{equation}
After that, we obtain the reconstructed sequence \TY{$\hat z^{1:t}$} by a transformer-based decoder:

\begin{equation}
\TY{\hat z^{1:t} = D(X^{1:T}),}
\end{equation}
\TY{where $D$ is the decoder ($D_{exp}$ or $D_{pose}$). } We train the encoder, decoder, and codebook via the loss function \cite{vqvae}:
\begin{equation}
\begin{split}
L_{vq} & \TY{= \|z^{1:t}- \hat z^{1:t}\|_2^2}       \\
   & \TY{+ \|sg[\hat X^{1:T}] -X^{1:T}\|_2^2 + \|sg[X^{1:T}] - \hat X^{1:T}\|_2^2,}
\end{split}
\end{equation}
where the first term represents the reconstruction loss of FLAME coefficients, and the latter two are employed to update the codebook by minimizing the distance between the codebook \TY{$C$} and the embedded features \TY{$\hat X^{1:T}$.} $sg[\cdot]$ represents a stop-gradient operation. 

The decoders and codebooks of the two VQ-VAEs are frozen and utilized in subsequent networks.

\begin{figure*}[t]
  \centering
    \includegraphics[width=1.0\linewidth]{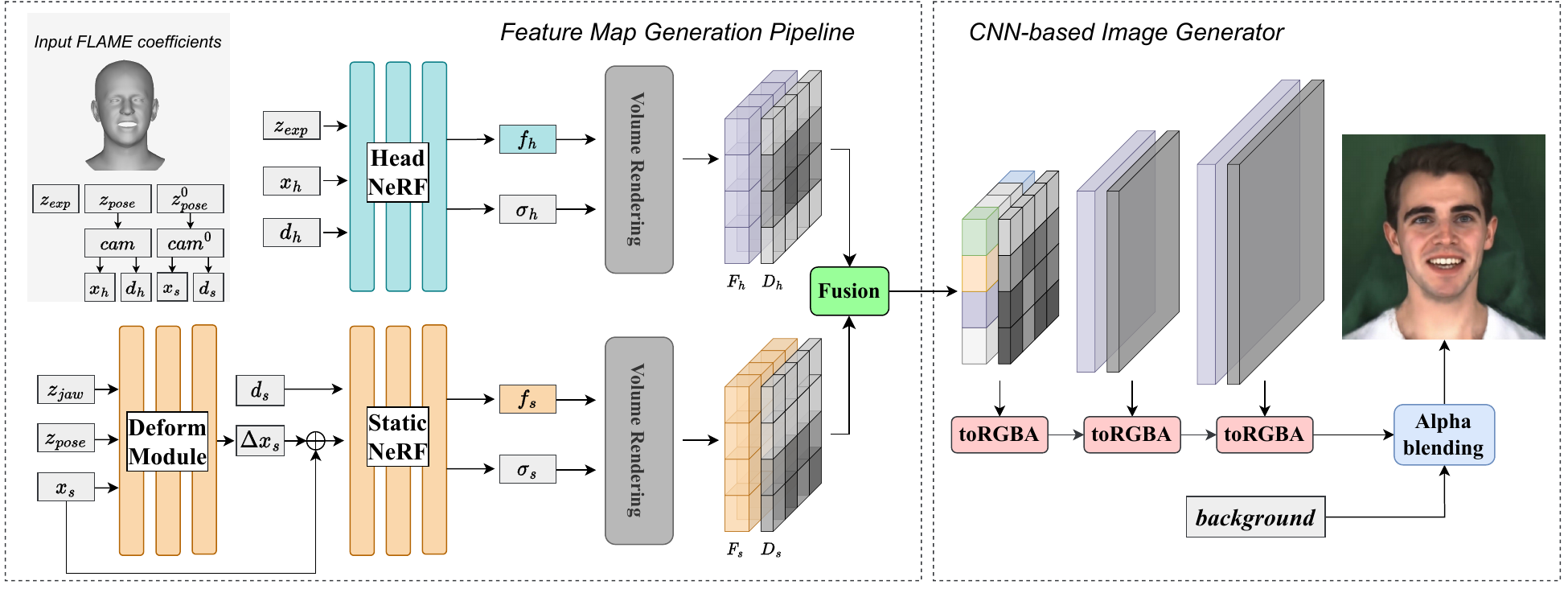}
  \caption{\TY{Overview of DBF-NeRF. Given expression $z_{exp}$, head pose $z_{pose}$, and initial head pose $z_{pose}^0$}, HeadNeRF and StaticNeRF generate two feature maps and density maps. Subsequently, the two sets of feature maps and density maps are fused using a density-based approach. Finally, after passing through a series of CNN upsampling layers, the ultimate high-resolution image is obtained.}
  \label{fig:expnerf}
\end{figure*}

\subsubsection{Audio to FLAME}
As shown in Fig.\ref{fig:a2f}. First, we employ two transformer-based style feature extractors to obtain the expression style feature $s_{exp}$ and the pose style feature $s_{p}$. Then, the style feature, conditional vector, and audio feature sequence are concatenated together as input sequences. Two new transformer-based decoders map the input sequences to the pre-trained expression codebook and pose codebook. \TY{After that}, we generate a style-controlled FLAME coefficients sequence that is synchronized with the speech utilizing frozen decoders. 
The implicit function of the ADF $h_a$ can be formulated as follows:

\begin{equation}
\TY{
     h_a:(A^{1:n}, V^{1:t}) \mapsto (\hat z_{exp}^{1: n},\hat z_{\Delta pose}^{1:n}),}
\end{equation}
where \TY{$A^{1: n}$ represents input audio of length $n$, $V^{1:t}$ represents style reference video of length $t$, $\hat z_{exp}^{1:n}$ and $\hat z_{\Delta pose}^{1:n}$ represent output expression sequence and head pose variation sequence}. \TY{During the training phase, the audio is extracted from the style video, hence $t=n$. During the testing, $n$ and $t$ may differ.}

\subsubsection{Loss}

(1) FLAME loss; our objective is to generate a FLAME coefficients sequence that closely resembles the ground truth as much as possible. This loss term can be formulated as follows.

\begin{equation}
    \TY{L_{f} = \sum_i \|\hat z_{exp}^i - z_{exp}^i\|^2_2 + \sum_i \|\hat z^i_{\Delta pose} - z^i_{\Delta pose}\|^2_2.}
\end{equation}

(2) GAN loss; for the sequence of generated expression coefficients and pose sequence, we employ a conditional discriminator to assess their authenticity. The discriminator takes the emotion label as a conditional input to determine the realism of the generated sequences. This loss term is formulated as:
\begin{equation}
\TY{
    L_{gan} = \arg \min_G \max_D (G,D).}
\end{equation}

(3) Contrastive loss; we employ a differentiable FLAME renderer to render the facial image through expression and pose coefficients. Then, we crop the rendered images and retain only the mouth region.  Two pre-trained encoders are employed to encode the cropped images and audio separately. We aim to minimize the difference between two encoded features. This loss term is formulated as:

\begin{equation}
\TY{
    L_{c} = \dfrac{e_m \cdot e_a}{\|e_m\|_2 \cdot \|e_a\|_2+\alpha}.
    }
\end{equation}

The total loss function can be formulated as: 

\begin{equation}
\TY{
    L_{total} =  L_{f} + L_{gan} +  L_{c}.}
\end{equation}

\subsection{DBF-NeRF}
In the context of the talk head generation task, the head and torso exhibit disparate motion patterns. In AD-NeRF \cite{adnerf}, the first step involves rendering the head image using Head-NeRF, followed by using the obtained image as a background to render the torso using Torso-NeRF. However, we have observed that this approach is akin to overlaying a torso image onto the head image, resulting in ineffective handling of the neck junction, thereby causing artifacts.
To address this issue, we propose the \textit{dual-branch fusion NeRF} (DBF-NeRF) model. As shown in Fig.\ref{fig:expnerf}, 
\TY{to model head and static regions, we employ two separate NeRFs (Head-NeRF and Static-NeRF) and volume renderer to predict high-dimensional feature maps.}
We then devised a feature map fusion method that is tailored specifically for NeRF. Subsequently, a series of CNN upsampling layers are employed to generate high-fidelity images. Overall, DBF-NeRF consists of four components: NeRF-based feature predictor, volume renderer, feature fusion operator, and CNN-based image generator.

The DBF-NeRF does not utilize audio features as input. Instead, we employ FLAME coefficients \cite{flame}. Specifically, for each frame, DBF-NeRF takes three components as input: (1) expression coefficients $z_{exp} \in \mathbb{R}^{53}$. 
(2) head pose coefficients \TY{$z_{pose} \in \mathbb{R}^{6}$} including global rotation and face translation. (3) head pose of the \TY{first frame $z_{pose}^0 \in \mathbb{R}^{6}$}.


\subsubsection{NeRF-based feature predictor}
For a 3D sampling point $x \in \mathbb{R}^3$, instead of predicting the RGB value, we predict a high dimensional feature vector $f \in \mathbb{R}^{256}$. For Head-NeRF, we convert the head pose \TY{$z_{pose}$} into camera parameter \TY{$cam \in \mathbb R^{4\times 4}$} to supply multi-viewpoint information, thereby enhancing rendering quality.
Therefore, the implicit formula for the Head-NeRF can be expressed as:
\begin{equation}
\TY{
    h_{head}: (\gamma(x_h),z_{exp},d_h) \mapsto (\sigma_h,f_h),}
\end{equation}
where $x_h$ represents a 3D point from one ray. $d_h$ represents the viewing direction. Both $x_h$ and $d_h$ are computed based on \TY{$cam$}, $\gamma$ represents the positional encoding function of NeRF \cite{nerf}. \TY{$\sigma_h \in \mathbb R^1$} represents the density value. \TY{$f_h \in \mathbb R^{256}$ represents the predicted feature vector.}

Since the transformed camera parameter from the head pose was not applicable to the torso, we employed a different strategy for Static-NeRF. The camera parameter of the initial frame \TY{$cam^0 \in \mathbb R^{4\times 4}$} are used for Static-NeRF. Additionally, we employed a deform module to model the non-rigid motion of the torso region. 
The implicit formula for deform module can be expressed as:
\begin{equation}
\TY{
   h_{deform}:(\gamma(x_s),\gamma(z_{pose}),\gamma(z_{jaw})) \mapsto \Delta x,}
\end{equation}
where \TY{$x_s$} is computed based on \TY{$cam^0$}, \TY{$z_{jaw} \in \mathbb R^3$} represents jaw rotation.
The implicit formula for the Static-NeRF can be expressed as:
\begin{equation}
\TY{
    h_{static}:(\gamma(x_s+\Delta x_s),d_s) \mapsto (\sigma_s, f_s).}
\end{equation}
\TY{where $d_s$ represents the viewing direction of first frame. $f_s \in \mathbb R ^{256}$ represents the predicted feature vector of Static-NeRF. $\sigma_s \in \mathbb R^1$ represents density value.}

\subsubsection{Volume renderer}
The head pose coefficients \TY{$z_{pose}$} is converted into camera parameter to generate rays $R \in \mathbb R^{n_r \times 6}$ on the low-resolution feature map, where $n_r$ represents the number of rays.
 
To obtain a pixel feature vector in the feature map, we use volume rendering by accumulating the sampled density feature value along the ray $r(t) = o+td$, where $o \in \mathbb R^3$ represents the ray emitted from the camera center $o$, $d \in \mathbb R^3$ represents the viewing direction. The final output feature $F(r)$ of one pixel can be evaluated as

\begin{equation}
    \TY{F(r) = \int_{t_n}^{t_f}f_\theta(r(t),d)\sigma_\theta(r(t))T(t)dt,}
\end{equation}

\noindent where $f_\theta$ and $\sigma_\theta$ represent the output of the MLP-based implicit function $h$ ($h_{head}$ or $h_{static}$),  $t_n$ and $t_f$ are the closest and farthest distances from the camera center $o$. $T(t)$ is the accumulated transmittance along the ray from $t_n$ to $t$ :
\begin{equation}
  \TY{T(t) = exp (-\int_{t_n}^{t}\sigma_\theta(r(t))dt).}
\end{equation}

For each frame, Head-NeRF generates two feature maps: \TY{head feature map $F_h \in \mathbb{R}^{64\times 64 \times 256}$ and head density map $D_h \in \mathbb{R}^{64\times 64 \times 1}$.} 
\TY{Static-NeRF generates two feature maps: static feature map $F_s\in \mathbb{R}^{64\times 64 \times 256}$ and static density map $D_s \in \mathbb{R}^{64\times 64 \times 1}$.}

\subsubsection{Feature fusion}
 
The part of the fusion of features takes \TY{$F_h, D_h, F_s, D_s$} as inputs. Using a density-based feature fusion method, we merged the head and static feature maps to obtain global features \TY{$F_{g}$} and global density \TY{$D_{g}$}. The density-based feature fusion function can be expressed as

\begin{gather}
    \TY{fusion(F_h,D_h,F_s,D_s) = \dfrac{(M_h\odot F_h \odot D_h + F_s \odot D_s)}{D_h + D_s + \alpha},} \notag \\
    \TY{F_g = fusion(F_h,D_h,F_s,D_s),} \\
    \TY{D_g = fusion(D_h,D_h,D_s,D_s),}\notag
\end{gather}    
where $M_h$ is a mask of the head region. \TY{$\alpha$ is a small value employed to prevent division by zero. $\odot$ represents Hadamard product. }

\begin{table*}[t]
\caption{Quantitative evaluation}
\label{quantitative}
\centering
\begin{adjustbox}{height=14.5mm}
\begin{tabular}{c|ccccc|ccccc}
\hline
          & \multicolumn{5}{c|}{MEAD}                                   & \multicolumn{5}{c}{LDST}                                                           \\
Method    & SSIM↑          & CPBD↑          & SyncNet↑       & M-LMD↓         & F-LMD↓         & SSIM↑          & CPBD↑          & SyncNet↑       & M-LMD↓         & F-LMD↓         \\ \hline
Wav2Lip   & 0.772          & 0.197          & \textbf{5.131} & \textbf{4.496} & \textbf{3.176} & 0.851          & 0.472          & \textbf{4.887} & 3.214          & 3.741          \\
PC-AVS    & 0.468          & 0.083          & 2.579          & 5.569          & 5.997          & 0.492          & 0.127          & 2.904          & 4.984          & 4.285          \\
EAMM      & 0.341          & 0.101          & 1.834          & 7.027          & 7.143          & 0.416          & 0.217          & 1.964          & 6.41           & 7.951          \\
SadTalker & 0.808          & 0.235          & 2.586          & 4.911          & 4.759          & 0.834          & 0.454          & 2.835          & 3.962          & 3.696          \\
AD-NeRF   & 0.804          & 0.188          & 3.302          & 4.554          & 4.213          & 0.82           & 0.434          & 3.049          & 3.253          & 3.684          \\
Ours      & \textbf{0.879} & \textbf{0.247} & 3.432          & 4.346          & 4.011          & \textbf{0.881} & \textbf{0.498} & 3.163          & \textbf{3.109} & \textbf{3.451} \\
GT        & 1              & 0.263          & 4.295          & 0              & 0              & 1              & 0.583          & 3.962          & 0              & 0              \\ \hline
\end{tabular}
\end{adjustbox}
\end{table*}

\subsubsection{CNN-based image generator}
To Get the final image, we employ a series of CNN upsampling layers. The upsampling layer function can be expressed as:
\begin{equation}
\begin{split}
    & \TY{U_i(X) = Conv2D(Y,K),} \\
    & \TY{Y = PixelShuffle(repeat(X,4) + \phi(X), 2),}
\end{split}
\end{equation}

\noindent where $\phi$ represents two Conv2D layers with $1\times 1$ kernel.
For a feature map input $X \in\mathbb R^{n\times n\times d}$, the repeated feature map $repeat(X,4) \in \mathbb R^{n\times n\times 4d} $ and $ \phi(X) \in \mathbb R^{n\times n\times 4d}$ are aggregated. Then the aggregated feature map is upsampled to $\mathbb R^{2n\times 2n\times d}$ by PixelShuffle \cite{pixelshuffle}. The final upsampled feature map is obtained by a Conv2D with K-kernel fix blur.

We map feature maps and density maps to RGBA instead of RGB by Conv2D with $1\times 1$ kernel and finally add the background through alpha blending.
 
\subsubsection{Loss}

(1) Photometric loss. For each image, our objective is to ensure that the RGB values of every pixel are consistent with the corresponding values in the real image. This loss term is formulated as:
\begin{equation}
\TY{
    L_{pho} = \|(\beta_hM_h+ \beta_sM_s)\odot(R-I_{GT}) \|^2_2,}
\end{equation}
where $R$ is the rendered image, $I_{GT}$ is the ground truth. $M_h$ and $M_s$ are masks for the head and static regions. $\beta$ represent weights, $\odot$ represents the Hadamard product operator.

(2) Perceptual loss. Unlike some previous works \cite{adnerf,dfrf,sspnerf}, we directly render the final image. Therefore, perceptual loss can be employed to effectively supervise network training. This loss term is formulated as:
\begin{equation}
    \TY{L_{per} = \sum_i \| \phi_i(R)-\phi_i(I_{GT}) \|^2_2,}
\end{equation}
where $\phi_i$ represent the i-th layer in VGG16 network.

The total loss can be formulated as:
\begin{equation}
    \TY{L_{total} = L_{pho}+L_{per}.}
\end{equation}

Compared to other NeRF-based approaches, DBF-NeRF has three advantages: (1) It fully leverages the NeRF capacity to represent three-dimensional scenes. (2) By fusing the features in the feature layer, the results generated around the neck area appear more natural. (3) It allows better utilization of perceptual loss in the supervision of network training.

\begin{figure*}
  \centering
    \includegraphics[width=0.9\linewidth]{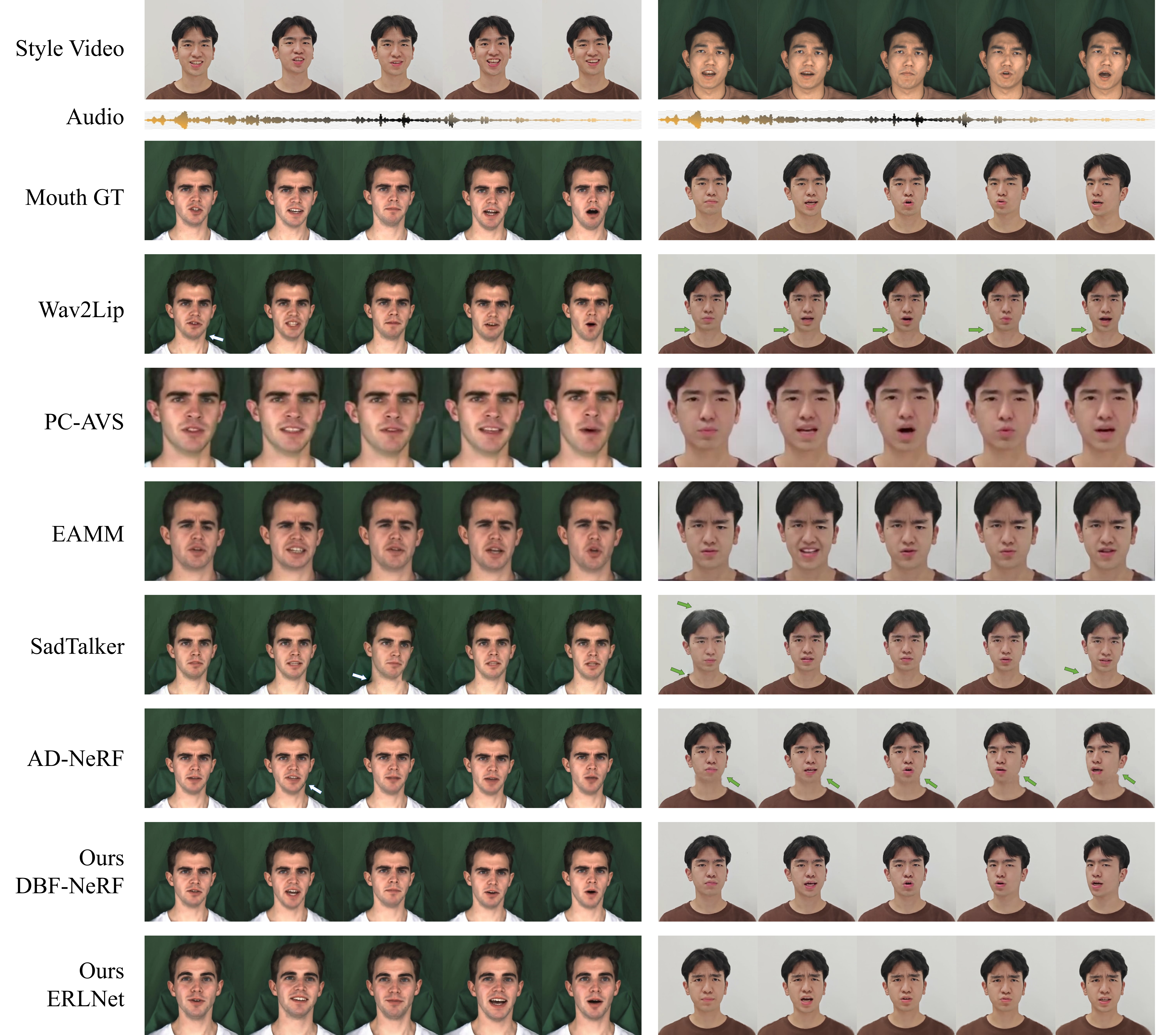}
    
  \caption{Qualitative Evaluation. We selected two style reference videos from the LDST and MEAD datasets and chose two subjects for our experiments. The first row corresponds to the style reference video, while the third row corresponds to the ground truth of mouth movements. The penultimate line corresponds to the results obtained by DBF-NeRF, while the last line represents the results generated by the complete ERLNet.
}
  \label{fig:qualitative}
\end{figure*}



\section{EXPERIMENTS}
\label{sec:experiment}

\subsection{Dataset}

To train and evaluate ERLNet, we use three datasets, LDST, MEAD, and HDTF. We created a new dataset named long-duration styled talking video (LDST), which contains 3 subjects with 5 different emotions, each video having a minimum duration of 30 seconds. There are no abrupt changes in the torso between different video segments of the same subject. MEAD is an emotional talking head dataset that contains 60 subjects with 8 different emotions and 3 different intensity levels. HDTF is a high-resolution in-the-wild audio-visual dataset. 

\subsection{Data Pre-processing}
First, we adjust the style reference video to a standardized format. This involves resizing the video to a resolution of 512x512 pixels and adjusting the frame rate to 25 FPS. Additionally, we ensure that the neck is positioned at the center of the video frame in the initial frame. Subsequently, we split the adjusted video into a sequence of individual images. For the input style reference video, our objective is to extract only the facial expressions and head pose information, while disregarding individual-specific details such as shape and texture. We used the pre-trained EMOCA \cite{EMOCA} model to predict the facial expression coefficients and pose coefficients for each image in the sequence.  We found that the predicted sequence of head pose is jittered, so a filter is used to smooth the head pose sequence. For the reference audio of the input content, we used the pre-trained transformer-based model Wav2Vec2\cite{wav2vec2} to extract the audio feature.
\begin{table}[t]
\caption{User Study}
\centering
\label{tab_userstudy}
\renewcommand\arraystretch{1}
\begin{tabular}{ccccc}
\hline
Method    & Quality& Expression    & Pose          & LipSync       \\ \hline
Wav2Lip   & 3.10                           & 4.53          & 2.47          & \textbf{4.07} \\
PC-AVS    & 2.80                           & 3.37          & 4.10          & 3.53          \\
AD-NeRF   & 3.60                           & 4.70          & 4.13          & 3.03          \\
EAMM      & 1.77                           & 1.07          & 3.27          & 3.43          \\
SadTalker & 3.93                           & 4.13          & 3.70          & 2.77          \\
Ours      & \textbf{4.37}                  & \textbf{4.73} & \textbf{4.27} & 3.87          \\ \hline
\end{tabular}
\end{table}

\subsection{Implementation Details}
We implement our framework in PyTorch \cite{pytorch}. \TY{We performed all experiments with a RTX 3090 GPU.} In our method, FLAME latent space, ADF and DBF-NeRF are individually trained. To train FLAME latent space, we employ AdamW \cite{adamw} optimizer ($\beta_1=0.9, \beta_2=0.999$ and $\epsilon=1e-8$), learning rate is initialized as $1e-4$. \TY{It took about 15 hours.} To train ADF and DBF-NeRF, we employ the Adam optimizer \cite{adam} with an initial learning rate $2e-5$ and $1e-4$.  \TY{It took about 3 days. In the volume rendering process, 64 points are sampled for each ray.} \TY{ During the inference phase, the generation of a single image with the resolution of $512 \times 512$ took about 1 second.}

\subsection{Quantitative Evaluation}
\subsubsection{Evaluation Metrics}
We use a variety of metrics to evaluate the performance of our model. To evaluate the quality of videos, we adopted the Structural Similarity Index Measure (SSIM) and the Cumulative Probability of Blur Detection (CPBD) metrics. To evaluate lip synchronization \cite{wav2lip}, we adopt the SyncNet confidence score and the landmark error of the mouth (M-LMD). To evaluate the quality of expressions, we adopt the facial landmark error (F-LMD). 


\subsubsection{Comparison with other methods}
We compared our method with some previous SOTA methods, including Wav2Lip \cite{wav2lip}, PC-AVS \cite{pcavs}, EAMM \cite{eamm}, AD-NeRF \cite{adnerf} and SadTalker \cite{sadtalker}.
We are in a self-driven setup to perform quantitative experiments on MEAD and LDST datasets. 
As shown in Tab.\ref{quantitative}, on the MEAD dataset, our approach yielded the best results in terms of SSIM and CPBD. In terms of SyncNet scores and landmark error metrics, only Wav2Lip outperformed our approach. This is because Wav2Lip generates lip sync based on the driving video, resulting in very low landmark errors. Additionally, Wav2lip benefits from training with SyncNet, which can lead to SyncNet scores even surpassing Ground Truth. On the LDST dataset, our approach yielded the best results in terms of SSIM, CPBD, M-LMD, and F-LMD. This implies that our ADF module is capable of generating realistic FLAME coefficients, and when DBF-NeRF receives these coefficients, it can generate high-quality images.

\subsection{Qualitative Evaluation}
In this experiment, we used the LDST and MEAD datasets. We compare our method with other methods including speaker-specific method (AD-NeRF) and speaker-agnostic methods (Wav2Lip, EAMM, PC-AVS and SadTalker). As shown in Fig.\ref{fig:qualitative}, our approach produces images of higher quality and is able to generate a result with natural expression and head pose. Among these methods, Wav2lip has the best lip synchronization, but there is a noticeable blurry region around the mouth. PC-AVS can control the head pose. EAMM can control expression, but in our experimental results we found that the results of eye generation are very strange. The results generated from EAMM and PC-AVS deviate significantly from the reference identity images. SadTalker exhibits limited diversity in lip variations, and its inability to handle torso motion results in a noticeable disconnection between the head and torso when the head pose changes.
AD-NeRF is also a method that uses NeRF as a renderer, but when the head movement is a little bit large, it leads to bad generation results at the connection between the head and the neck. Furthermore, AD-NeRF is tightly coupled to audio, making it impossible to facilitate style control. 
Benefiting from the feature map fusion mechanism of DBF-NeRF, our method
eliminates
the occurrence of artifacts around the neck region.

\subsection{User Study}
We conducted a user study to further evaluate our model. We invited 30 volunteers, and each of the volunteers was asked to score video quality, realism of the style (expression changes and head movements), and lip synchronization of 30 video clips. The videos were generated using the LDST and MEAD testset. We integrated and averaged the collected data. As shown in Tab.\ref{tab_userstudy}. Our approach achieves the most realistic results.


\begin{figure}[]
  \centering
   \includegraphics[width=1.0\linewidth]{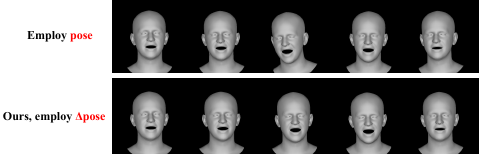}
   \caption{Ablation study of pose representation.}
   \label{fig:ablation_pose}
\end{figure}

\subsection{Ablation Study}


\begin{table}[]
\caption{Ablation Study on the MEAD}
\centering
\label{tab_ablation}
\begin{tabular}{@{\hspace{3pt}}c@{\hspace{7pt}}c@{\hspace{7pt}}c@{\hspace{7pt}}c@{\hspace{7pt}}c@{\hspace{7pt}}c} 
\hline
Method   & SSIM↑          & CPBD↑          & SyncNet↑       & M-LMD↓         & F-LMD↓          \\ 
\hline
\makecell[c]{w/o Deform, $z_{pose}$}  & 0.851          & 0.212          & 3.343          & 4.352          & 4.092           \\
\makecell[c]{w/o Deform, $z_{pose}^0$}  & 0.833          & 0.203          & 3.359          & 4.347          & 4.133           \\
w/o Contrastive & 0.858          & 0.244          & 2.304          & 5.302          & 4.787           \\
$\Delta$expression     & 0.847          & 0.239          & 2.867          & 4.995          & 5.923           \\
Ours     & \textbf{0.879} & \textbf{0.247} & \textbf{3.432} & \textbf{4.346} & \textbf{4.011}  \\
\hline
\end{tabular}
\end{table}

\subsubsection{Forms of intermediate representation}
For the ADF module, we explored four different forms of representation for expression and head pose: (1) expression. (2) pose. (3) $\Delta$expression. (4) $\Delta$pose. As shown in Fig.\ref{fig:ablation_pose}, we observed that utilizing $\Delta$pose as the representation results in smoother pose sequences. On the contrary, as shown in Tab.\ref{tab_ablation}, employing the $\Delta$ expression as the representation leads to a decline in network performance. 
In our conjecture, this phenomenon can be attributed to the correspondence between specific audio features and facial expressions, including the shape of the mouth, with less influence from the preceding frame.


\subsubsection{Speech contrastive loss}
As shown in Tab.\ref{tab_ablation} we directly removed speech contrastive loss, leading to a significant decrease in lip synchronization. Without contrastive speech loss, the ADF module struggles to capture mouth movements effectively and tends to pay more attention to facial expressions.

\begin{figure}[]
  \centering
   \includegraphics[width=1.0\linewidth]{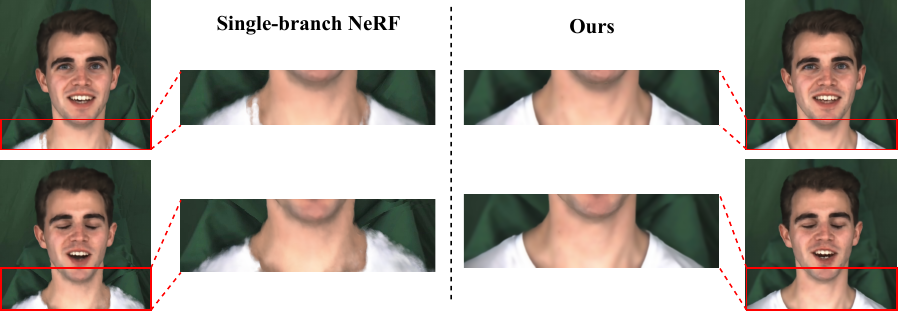}
   \caption{\TY {Ablation study of DBF-NeRF indicates that employing a single-branch NeRF leads to a significant deterioration in rendering performance on the torso.} }
   \label{fig:ablation-fusion}
\end{figure}

\subsubsection{DBF-NeRF} 
\TY{We removed the dual-branch fusion module and rendered the entire image using a single NeRF channel. As shown in Fig.\ref{fig:ablation-fusion}, the use of a single-branch NeRF leads to a significant deterioration in the performance of the torso.}

\subsubsection{Deform module} 
\begin{figure}[]
  \centering
   \includegraphics[width=1.0\linewidth]{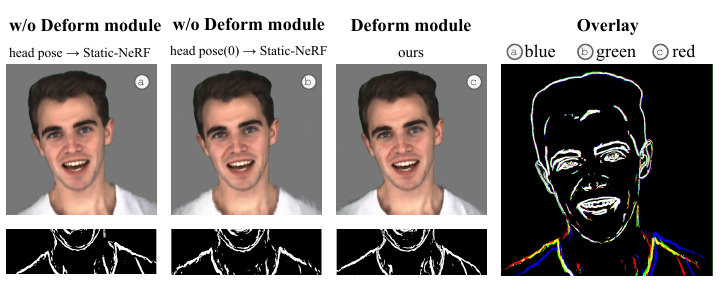}
   \caption{Ablation study of deform module. \TY{(a) Incorrect torso angle, 
   highlighted by the blue line in the overlaid map. 
   (b) Blurred edges. (c) Correct torso angle with clear edges.} }
   \label{fig:ablation-deform}
\end{figure}

As shown in Fig.\ref{fig:ablation-deform}, we evaluate our model under three settings. (a) w/o deform module, for Static-NeRF, employ head pose as camera parameter. This would result in the torso following the motion of the head. This phenomenon becomes particularly pronounced when there is a substantial degree of head movement. (b) w/o deform module, for Static-NeRF, employ head pose from the first frame as camera parameter. This approach does not fully harness the three-dimensional representational capabilities of NeRF, which consequently results in a degradation in rendering quality. (c) Ours employs a deform module to wrap Static-NeRF. We employed the deform module to distort sampled rays, enabling the feature maps generated by Static-NeRF to align with head movement. Our approach achieves a more stable torso while maintaining a high rendering quality.

\subsubsection{Perceptual loss} 
\begin{figure}[]
  \centering
   \includegraphics[width=1.0\linewidth]{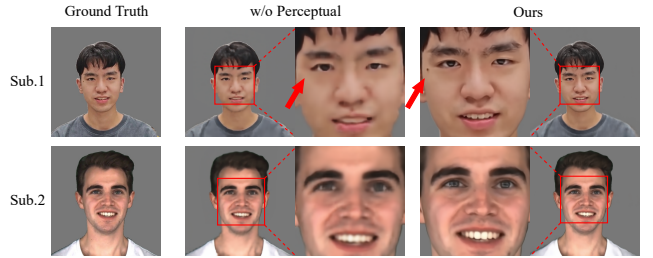}
   \caption{Ablation study of VGG16 perceptual loss. \TY{Without VGG16 perceptual loss, the generated results lack facial details, such as the mole on Sub1. Readers are encouraged to zoom in for optimal visualization.}}
   \label{fig:ablation_vgg}
\end{figure}

Our network utilizes NeRF to generate low-resolution features, allowing the generation of an entire image in a single iteration. As shown in Fig.\ref{fig:ablation_vgg}, we directly removed the perceptual loss of VGG16. With the same number of iterations, using only the photometric loss resulted in a blurrier generation effect, making it challenging to capture high-frequency facial details.

\section{DISCUSSION}

\TY{ERLNet has demonstrated remarkable capabilities in the field of talking head video generation. However, our method still has limitations. ERLNet can only render partial upper-body images and cannot capture actions involving arm movements. Another limitation is that our model cannot specify the observer's viewpoint. Enabling free-viewpoint speech-driven talking head video generation is an interesting topic. We will address these issues in future work.}


\section{CONCLUSION}
In summary, we propose ERLNet, a style-controllable talking head video generation method based on NeRF. Divergent from previous approaches, our methodology employs FLAME coefficients as an intermediate representation instead of relying on audio features. We propose the ADF module for the synchronized generation of stylized FLAME coefficients with audio inputs. Furthermore, a dual-branch fusion NeRF structure is proposed to enhance rendering quality. Additionally, we collected a dataset of talking head videos that encompass various emotional expressions named LDST. Compared to previous state-of-the-art methods, our approach demonstrates the ability to produce higher-quality images while also learning distinct expression styles and head pose styles, thereby enhancing the realism of our generated results.

{\small
\bibliographystyle{ieee}
\bibliography{egbib}
}

\end{document}